\documentclass[conference]{IEEEtran}
\IEEEoverridecommandlockouts

\usepackage[hidelinks]{hyperref}

\usepackage{orcidlink}
\usepackage{amsmath,amssymb,amsfonts}
\usepackage{algorithmic}
\usepackage[linesnumbered,ruled,vlined]{algorithm2e}
\usepackage{graphicx}
\usepackage{subcaption}
\usepackage{textcomp}
\usepackage{multirow}
\usepackage{balance}
\usepackage{booktabs}
\usepackage{orcidlink}
\usepackage[export]{adjustbox}

\usepackage{subcaption}
\usepackage{textcomp}
\usepackage{balance}
\usepackage{url}
\usepackage{caption}

\usepackage{authblk}

\def\BibTeX{{\rm B\kern-.05em{\sc i\kern-.025em b}\kern-.08em
    T\kern-.1667em\lower.7ex\hbox{E}\kern-.125emX}}
\newcommand{\linebreakand}{%
  \end{@IEEEauthorhalign}
  \hfill\mbox{}\par
  \mbox{}\hfill\begin{@IEEEauthorhalign}
}

\def\BibTeX{{\rm B\kern-.05em{\sc i\kern-.025em b}\kern-.08em
    T\kern-.1667em\lower.7ex\hbox{E}\kern-.125emX}}

\usepackage[
  backend=biber,
  style=ieee,
  minnames=1,
  maxcitenames=2, maxbibnames=6
]{biblatex}

\begin{document}


\title{VisionGuard: Synergistic Framework for Helmet Violation Detection}

\author[1, 2]{Lam-Huy Nguyen\orcidlink{0009-0003-2890-5741}*\thanks{*Equal contributions.}}
\author[1, 2]{Thinh-Phuc Nguyen\orcidlink{0009-0000-5150-9307}*}
\author[1, 2]{Thanh-Hai Nguyen\orcidlink{0009-0004-7953-8942}*}
\author[1, 2]{Gia-Huy Dinh\orcidlink{0009-0004-8746-3004}*}
\author[1, 2]{\authorcr Minh-Triet Tran\orcidlink{0000-0003-3046-3041}}
\author[1, 2]{Trung-Nghia Le\orcidlink{0000-0002-7363-2610}**\thanks{**Corresponding author. {\it e-mail: ltnghia@fit.hcmus.edu.vn}}}

\affil[1]{University of Science, Ho Chi Minh City, Vietnam}
\affil[2]{Vietnam National University, Ho Chi Minh City, Vietnam}



\maketitle

\begin{abstract}

Enforcing helmet regulations among motorcyclists is essential for enhancing road safety and ensuring the effectiveness of traffic management systems. However, automatic detection of helmet violations faces significant challenges due to environmental variability, camera angles, and inconsistencies in the data. These factors hinder reliable detection of motorcycles and riders and disrupt consistent object classification. To address these challenges, we propose VisionGuard, a synergistic multi-stage framework designed to overcome the limitations of frame-wise detectors, especially in scenarios with class imbalance and inconsistent annotations. VisionGuard integrates two key components: Adaptive Labeling and Contextual Expander modules. The Adaptive Labeling module is a tracking-based refinement technique that enhances classification consistency by leveraging a tracking algorithm to assign persistent labels across frames and correct misclassifications. The Contextual Expander module improves recall for underrepresented classes by generating virtual bounding boxes with appropriate confidence scores, effectively addressing the impact of data imbalance. Experimental results show that VisionGuard improves overall mAP by 3.1\% compared to baseline detectors, demonstrating its effectiveness and potential for real-world deployment in traffic surveillance systems, ultimately promoting safety and regulatory compliance.
\end{abstract}

\section{Introduction}



Helmet rule violation detection is a critical component of road safety enforcement, particularly in regions where motorcycles serve as a primary mode of transportation. The demand for effective traffic surveillance systems is especially pronounced in many developing Asian countries, where traffic infrastructure, public safety regulations, and enforcement mechanisms are often underdeveloped \cite{wismans2016commentary}. Among the most frequently violated traffic regulations in Southeast Asia are motorcycle helmet laws \cite{peltzer2014helmet}. Implementing automated helmet violation detection systems can support law enforcement by identifying offenders and enabling timely penalties, ultimately encouraging behavioral change among commuters and improving public safety \cite{alimohamadi2024enhancing}. As such, developing a reliable and efficient automatic detection system for motorcycle helmet violations is of vital importance.

Object detection forms the backbone of intelligent traffic monitoring and autonomous penalty enforcement systems. Numerous methods have been proposed for this task. Deep learning-based detectors such as R-CNNs~\cite{RCNN} and the YOLO family of models~\cite{du2018understanding, terven2023comprehensive, hussain2023yolo} have demonstrated strong performance in terms of accuracy and robustness, with YOLO being particularly well-suited for real-time applications. More recent transformer-based models, such as DETR~\cite{carion2020end}, Deformable DETR~\cite{zhu2020deformable}, and Swin Transformer~\cite{liu2021swin}, show promise in handling complex visual conditions like occlusion, but their high computational demands often make them unsuitable for real-time deployment in practical surveillance settings.

\begin{figure}[t!]
  \centering
  \begin{subfigure}[t]{0.49\linewidth}
    \centering
    \includegraphics[width=\linewidth, height=5cm]{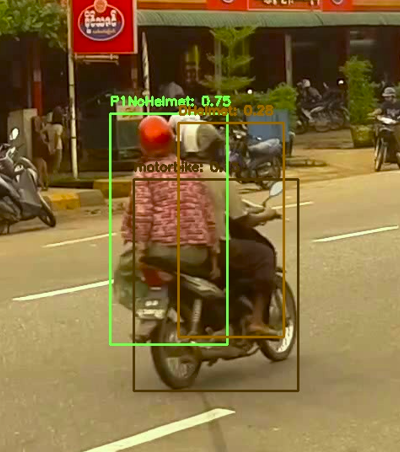}
    \caption{Viewed from behind}
    \label{fig:Occlusion}
  \end{subfigure}
  \hfill
  \begin{subfigure}[t]{0.49\linewidth}
    \centering
    \includegraphics[width=\linewidth, height=5cm]{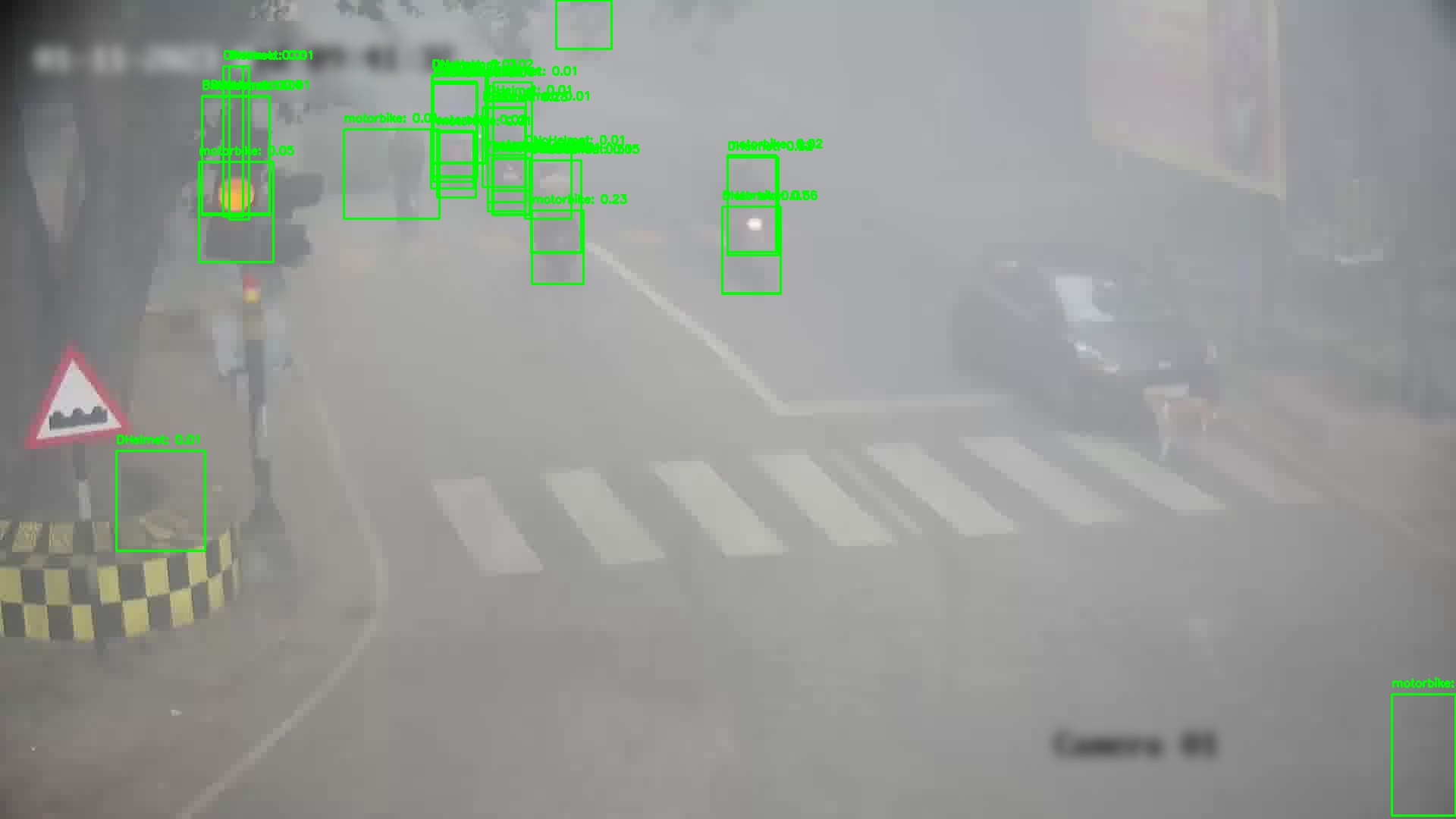}
    \caption{Heavy smog}
    \label{fig:Overlapping}
  \end{subfigure}
  \caption{Examples of challenges in helmet violation detection.}
  \label{fig:combined_occlusion_smog}
\end{figure}

Despite these advancements, vision-based detection systems continue to face limitations stemming from physical camera constraints. Surveillance cameras are typically installed at elevated positions, resulting in oblique or top-down views that obscure critical visual details, especially in scenarios involving multiple passengers on a single motorcycle (\autoref{fig:Occlusion}). This occlusion problem is exacerbated in densely populated urban environments with heavy traffic, where accurately identifying the violator becomes increasingly difficult. Furthermore, adverse weather conditions such as rain, smog, or poor lighting significantly degrade image quality and hinder the performance of existing detection models (\autoref{fig:Overlapping}). These challenges highlight the need for a robust, real-time detection framework that can perform reliably in diverse and uncontrolled environments. 


\begin{figure*}[t!]
    \centering
    \includegraphics[width=\textwidth]{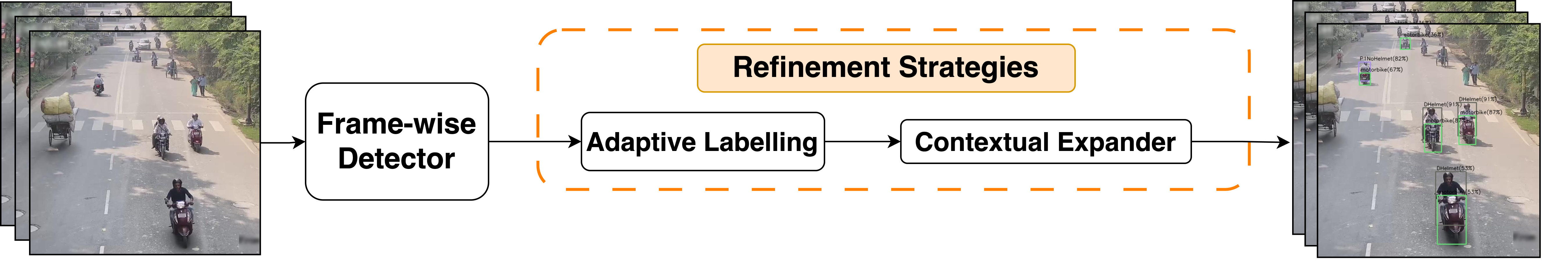}
    \caption{Overview of our proposed VisionGuard framework which consists of a frame-wise detector whose results are refined through Adaptive Labeling and Contextual Expander modules.}
    \label{fig:MethodPipeline}
\end{figure*}

To address the limitations of existing object detection methods, we propose VisionGuard, a novel synergistic multi-stage framework designed to enhance detection performance. VisionGuard integrates post-processing strategies to refine the raw outputs from detectors, thereby improving overall system accuracy. We introduce Adaptive Labeling, a tracking-based refinement module that enhances classification consistency by leveraging the OC-SORT tracking algorithm~\cite{cao2023observationcentricsortrethinkingsort}. This ensures that each detected instance retains a consistent label across frames, enabling the correction of misclassifications. Additionally, we present the Contextual Expander module to improve recall for underrepresented classes. This module generates virtual bounding boxes for relevant detections, assigning hierarchical confidence scores that prioritize rare classes, ultimately enhancing the detection of underrepresented objects. 

We conducted experiments using the AI City Challenge 2023 and 2024 dataset~\cite{aicity}. Our proposed framework was applied to two state-of-the-art frame-wise detectors: the end-to-end DETR model~\cite{carion2020end} and the ensembled Co-DETR models~\cite{zong2023detrs}. Both detectors were selected for their robustness in handling complex detection tasks in traffic surveillance. The application of our proposed framework led to a 3.1\% improvement in the overall mAP score compared to the baseline detectors. This demonstrates the effectiveness of VisionGuard in refining detection results and enhancing classification consistency, especially in the presence of class imbalance and inconsistent annotations. It also shows potential of VisionGuard for deployment in traffic management systems, promoting safety and regulatory compliance through improved violation detection. 


Our main contributions are as follows: 
\begin{itemize}
    \item We propose a novel synergistic multi-stage framework to address the existing challenges of state-of-the-art frame-wise detectors. 
    \item We introduce the tracking-based refinement Adaptive Labeling module to mitigate class switching of objects by averaging confidence scores from different frames. 
    \item We present the the Contextual Expander module to enhance the prediction of rare classes by adding suitable virtual bounding boxes with reasonable confidence score.
\end{itemize}

\section{Related Work}

\subsection{Helmet Rules Violation Detection:} 

Safety in transportation is a crucial criterion to which all authorities pay attention. Such problem has become a track in AI City Challenge \cite{aicity}, with various studies aiming to detect helmet rule infractions.

YOLO-based models~\cite{du2018understanding} \cite{terven2023comprehensive} \cite{hussain2023yolo} were widely used for their speed and accuracy. Tsai et al. focused on detecting helmet violations using YOLOv7-E6E \cite{wang2023yolov7} as a baseline, proposing YOLOv7-CBAM \cite{yue2023environmental} and YOLOv7-SimAM \cite{ning2024yolov7} for improved performance. Aboah et al. used YOLOv8 \cite{swathi2024yolov8} and few-shot data sampling to develop a robust real-time helmet detection model \cite{aboah2023real}, balancing accuracy and performance for real-world applications.

Transformer-based networks also showed promise in helmet detection. Chen et al. fused Co-DETR \cite{zong2023detrs} and DETA \cite{yang2022deta} models for improved detection \cite{aneff}, while Cui et al. used ensemble modeling with DETA to address category imbalance and refine bounding boxes with the Passenger Recall Module (PRM) \cite{cui2023effective}. The SORT algorithm \cite{bewley2016simple} \cite{abouelyazid2023comparative} helped minimize category switching during movement. Hao Vo et al. \cite{vo2024robust} used ensemble modeling and post-processing techniques like Weighted Box Fusion and Minority Optimizer.

While existing approaches have improved detector performance through model architecture enhancements and post-processing techniques, they largely operate on a frame-wise basis, limiting temporal consistency and recall for rare classes. In contrast, our method introduces a multi-stage framework to systematically address classification inconsistency and recall degradation by leveraging temporal coherence and strategic bounding box generation, marking a significant advancement over prior methods.

\subsection{Detection-based Multiple Object Tracking}

This approach uses object detections to initialize and update trajectories, separating detection and tracking. Detectors locate objects, and algorithms associate detections across frames, maintaining consistent identities through occlusions or motion. Methods like SORT~\cite{bewley2016simple}, DeepSORT~\cite{wojke2017simple}, and OC-SORT~\cite{cao2023observationcentricsortrethinkingsort} combine Kalman filtering for motion prediction with the Hungarian algorithm for data association. SORT uses predicted boxes, while DeepSORT adds appearance features to handle long-term occlusions. Recent variants like OC-SORT~\cite{cao2023observationcentricsortrethinkingsort} and BoT-SORT~\cite{aharon2022bot} improve performance in crowded scenes. These methods balance accuracy and real-time performance.

In this paper, we use OC-SORT~\cite{cao2023observationcentricsortrethinkingsort} as the primary tracker before the refinement step, ensuring consistency in detecting instances, particularly when multiple motorbikes and passengers appear in the same frame. 




\section{Proposed Method}
\label{Methodology}

\subsection{Overview}

This paper aims to improve helmet rule violation detection by integrating targeted refinement strategies into the detection pipeline. As illustrated in \autoref{fig:MethodPipeline}, our method begins with frame-wise object detection using state-of-the-art detectors such as Co-DETR~\cite{zong2023detrs} and DETR~\cite{carion2020end}. Eventually, we apply tracking to maintain temporal consistency across frames and then adjust the confidence scores for improved  robustness and precision of motorcycle helmet rule violation detection in surveillance cameras.
 



\subsection{Tracking-based Adaptive Labeling}

\subsubsection{{Tracking with OC-SORT}}


\begin{figure}[t!]
    \centering
    \includegraphics[width=\linewidth]{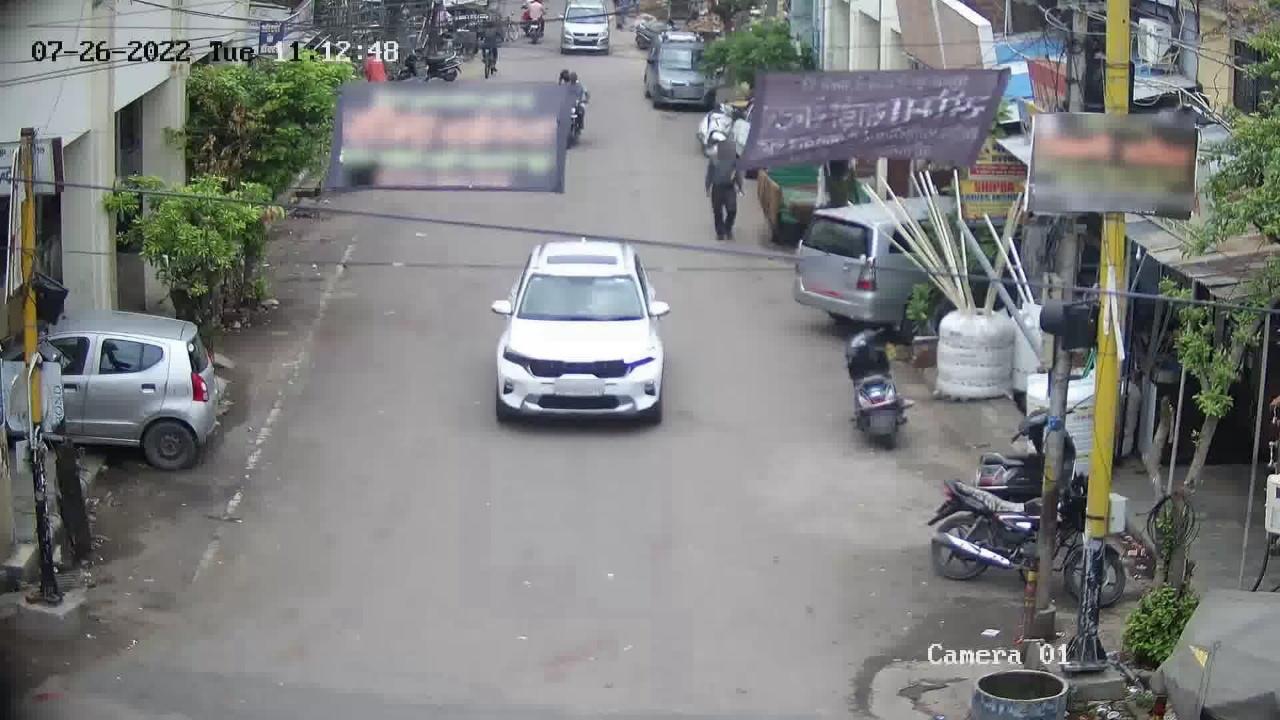}
    \caption{Occlusion due to high camera angle.}
    \label{fig:occlusion_high_angle}
\end{figure}

Frame-wise detectors often suffer from inconsistent classifications, where an object may be correctly labeled in most frames but misclassified in others. To address this, we adopt the assumption that an object's class should remain consistent over time and use tracking to refine misclassified instances. We employ OC-SORT~\cite{cao2023observationcentricsortrethinkingsort}, which leverages object observations rather than linear motion models to estimate trajectories, making it more robust to occlusions. This is particularly beneficial for helmet rule violation detection, where high-angle surveillance and dense traffic often lead to partial visibility and frequent occlusions (\autoref{fig:occlusion_high_angle}).

Each detected object is assigned an ID in the first frame and tracked across subsequent frames. OC-SORT updates each track by matching new detections with previous ones, preserving bounding box and class information. New objects are initialized with new IDs and added to separate tracks. These temporally consistent tracks are later used in the Adaptive Labeling module to correct classification errors and improve label stability across frames.

\subsubsection{Adaptive Labeling}
\label{subsubsection: AdaptiveLabelling}

We propose an adaptive refinement strategy that improves classification accuracy by leveraging temporal consistency and confidence stability across object tracks. Frame-wise detectors often produce inconsistent predictions for the same object across consecutive frames, particularly under occlusion or partial visibility. Our approach addresses these inconsistencies through a two-stage process: {Track Quality Assessment} and {Label Correction}. 

\paragraph{Track Quality Assessment} 
Given a track $t$ consisting of bounding boxes $\{b_1, b_2, \ldots, b_n\}$, where each detection $b_i$ has a predicted label $l_i$ and confidence score $c_i$, we define a track quality score $Q_t$ as:
\begin{equation}
    Q_t = (1 - r) \cdot \bar{c}_t,
\end{equation}
where $\bar{c}_t$ represents the average confidence over the entire track and $r$ denotes the label change ratio, computed as the proportion of consecutive frames where the predicted label differs. This metric captures both the temporal stability and average reliability of predictions in the track.

To assign a consistent label $L_t$ for the track, we use weighted voting where each label $l$ receives a vote equal to the sum of confidence scores from all detections with that label:
\begin{equation}
    L_t = \arg\max_l \sum_{i: l_i = l} c_i.
\end{equation}

Only tracks with $Q_t \geq \theta_q$, where $\theta_q$ is a fixed quality threshold, are considered for refinement.

\paragraph{Label Correction}
For each detection $b_i$ in a qualified track, we compute an adaptive threshold $\theta_i$ as follows:
\begin{equation}
    \theta_i = \left(\theta_0 + \alpha (1 - \bar{c}_t)\right) \cdot \left(1 + (0.5 - Q_t)\right),
\end{equation}
where $\theta_0$ is the base confidence threshold, and $\alpha$ controls the weight of the confidence penalty based on track reliability.

A detection is eligible for refinement if it meets the following criteria:

\begin{itemize}
    \item The predicted label $l_i$ deviates from the consistent track label $L_t$;
    \item The label $l_i$ does not belong to a protected set of high-confidence classes (e.g., $l_i \notin \{1, 2, 3\}$);
    \item The associated confidence score $c_i$ falls below the adaptive threshold $\theta_i$.
\end{itemize}

Prior to relabeling, we check for high spatial agreement with existing detections of class $L_t$. A match is valid if the Intersection-over-Union (IoU) exceeds 0.8 and the confidence score of the matching detection is sufficiently high. If no such match is found, the detection is relabeled to $L_t$ and its confidence is penalized:
\begin{equation}
    c_i = \lambda \cdot c_i,
\end{equation}
where $\lambda < 1$ is a fixed penalty factor.

If no valid spatial match is found and the bounding box area exceeds a minimum threshold, the detection is removed to suppress potential false positives.

This module adaptively refines object labels by leveraging temporal cues, confidence analysis, and spatial alignment. The result is a more stable and accurate classification across video frames, which is essential in real-world surveillance scenarios such as helmet violation detection.

\subsection{Contextual Expander}

\begin{figure}[t!] 
\centering
\includegraphics[width=\linewidth]{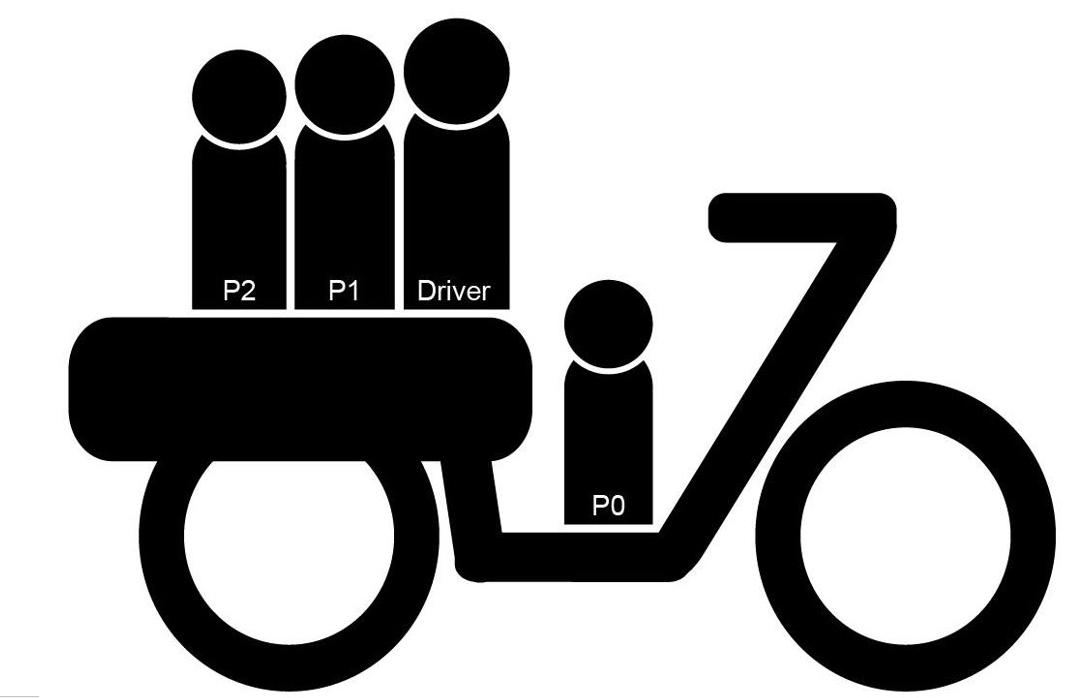} 
\caption{Illustration of typical rider positions on a motorcycle.} 
\label{fig:rider_position} 
\end{figure}

In practical traffic surveillance scenarios, motorcycles frequently carry multiple riders in complex seating arrangements. A common example includes a driver with three passengers ordered as P0 (front), Driver, P1 (middle), and P2 (rear), as illustrated in Fig.\ref{fig:rider_position}. Among these positions, P0 and P2 occur infrequently in real-world data, leading to limited training instances and reduced model confidence in detecting these classes. To address this, we introduce the Contextual Expander module. Drawing inspiration from recent work on enhancing detection robustness with virtual bounding boxes\cite{vo2024robust} and improving confidence scores for underrepresented classes~\cite{van2024motorcyclist}, this module generates context-aware virtual bounding boxes under predefined spatial constraints. It also assigns class-specific confidence scores based on empirical data distributions, enabling the system to simulate likely but low-confidence detections and improve recall for rare rider positions. The pseudocode for the Contextual Expander is provided in \autoref{alg:contextual_expander}.


\begin{table}[t!]
\centering
\caption{Labels used in the AI City Challenge 2024’s dataset.}
\resizebox{\linewidth}{!}{
\begin{tabular}{|c|c|c|}
\hline
\textbf{Label ID} & \textbf{Label Category} & \textbf{Description} \\ \hline
1 & motorbike      & Bounding box for motorbikes      \\ \hline
2 & DHelmet      & The driver is wearing helmet      \\ \hline
3 & DNoHelmet      & The driver is not wearing helmet      \\ \hline
4 & P1Helmet      & The P1 is wearing helmet      \\ \hline
5 & P1NoHelmet      & The P1 is not wearing helmet       \\ \hline
6 & P2Helmet     & The P2 is wearing helmet     \\ \hline
7 & P2NoHelmet      & The P2 is not wearing helmet      \\ \hline
8 & P0Helmet      & The P0 is wearing helmet      \\ \hline
9 & P0NoHelmet      & The P0 is not wearing helmet     \\ \hline
\end{tabular}
}

\label{tab:Table3}
\end{table}

\begin{algorithm}[t!]
\caption{Contextual Expander}
\label{alg:contextual_expander}
\KwIn{
    $motor\_list$, $human\_list$: Detected instances with \texttt{bbox}, \texttt{class}, \texttt{confidence} \\
    $class\_list$: Set of target classes, shown in \autoref{tab:Table3}
}
\KwOut{Augmented list of predicted instances $results$}

\textbf{Remove overlapping} instances with IoU $\geq 0.8$\;

\textbf{Initialize} $results \leftarrow \emptyset$\;

\ForEach{$motor \in motor\_list$}{
    $\texttt{results} \leftarrow \texttt{results} \cup \{\texttt{motor}\}$\;
    
    \ForEach{$cl \in class\_list$ \textbf{where} $cl \neq motor.class$}{
        $c \leftarrow 1 \times 10^{-5}$ \;
         $\texttt{virtual} \leftarrow \{ cl, motor.bbox, c \} $ \;
        $\texttt{results} \leftarrow \texttt{results} \cup \{\texttt{virtual}\}$\;
    }
}

\ForEach{$human \in human\_list$}{
    $\texttt{results} \leftarrow \texttt{results} \cup \{\texttt{human}\}$\;
    \ForEach{$cl \in class\_list$ \textbf{where} $cl \neq human.class$}{
        $c \leftarrow 1 \times 10^{-4}$\;
        \If{$cl \in \{4, 6, 7, 8, 9\}$}{
            $c \leftarrow c + 3 \times 10^{-5}$\;
        }

        \uIf{$cl = 1$ \textbf{and} $human.class \in \{2, 3\}$ \textbf{and} $human.conf > 0.01$}{
             $\texttt{virtual} \leftarrow \{ cl, human.bbox, c \} $ \;
            $\texttt{results} \leftarrow \texttt{results} \cup \{\texttt{virtual}\}$\;
        }
        \uElseIf{$cl = 9$ \textbf{and} $human.class \in \{2, 3\}$ \textbf{and} $human.conf > 0.1$}{
            $human.bbox \leftarrow human.bbox * 0.7$ \;
            $\texttt{virtual} \leftarrow \{ cl, human.bbox, c \} $ \;
            $\texttt{results} \leftarrow \texttt{results} \cup \{\texttt{virtual}\}$\;
        }
        \uElseIf{$cl = 7$ \textbf{and} $human.class = 5$ \textbf{and} $human.conf > 0.01$}{
            $\texttt{virtual} \leftarrow \{ cl, human.bbox, c \} $ \;
            $\texttt{results} \leftarrow \texttt{results} \cup \{\texttt{virtual}\}$\;
        }
        \uElseIf{$cl \in \{2, 3, 4, 5\}$}{
            $\texttt{virtual} \leftarrow \{ cl, human.bbox, c \} $ \;
            $\texttt{results} \leftarrow \texttt{results} \cup \{\texttt{virtual}\}$\;
        }
    }
}

\Return $results$\;
\end{algorithm}

The process starts by removing overlapping detections of the same class with an IoU above 0.8 to avoid conflicts. For each detected motorbike, synthetic instances of all other target classes are added at the same location with a low confidence score (0.00001), reflecting the low likelihood of significant overlap with ground truth annotations for those classes. For example, a driver overlapping with the motorbike instance at an IoU greater than 50\% is relatively low.

For each detected human instance, synthetic predictions for related classes are generated based on the source class and its confidence. For instance, if a driver (helmeted or not) is detected with confidence above 0.01, a synthetic prediction for a correlated class is added with a default confidence of 0.0001. This leverages observed co-occurrence patterns, such as the frequent presence of P0 or P1 near a detected driver. To better match real-world scale, we introduce contextual virtual bounding boxes for P0 by scaling the driver’s bounding box to 70\% of its original size.

Inspired by Luong et al.~\cite{van2024motorcyclist}, we address the underrepresentation of certain classes in top-ranked predictions by applying class-dependent confidence adjustments. Specifically, we boost the confidence scores of rare classes with small additive offsets. This increases the likelihood that these instances exceed the confidence threshold and are included among the top 100 detections per frame. Since mAP evaluation typically considers only the top 100 predictions ranked by confidence, this targeted adjustment improves the precision of rare classes and contributes to a higher overall mAP score. 

\section{Experiments}

\subsection{Implementation Details}

We evaluated our VisionGuard framework using two state-of-the-art detectors: DETR~\cite{carion2020end} and an ensemble of Co-DETRs~\cite{vo2024robust}.

DETR~\cite{carion2020end} eliminates the need for anchor boxes or region proposals by directly predicting a fixed number of objects through bipartite matching. We trained DETR with a ResNet-50 backbone for 22 epochs using the default configuration.

The Co-DETR ensemble~\cite{vo2024robust} combines predictions from multiple pre-trained Co-DETR checkpoints~\cite{zong2023detrs}, each operating at different input resolutions ($640 \times 640$ and $1280 \times 1280$). Final outputs are merged using Weighted Box Fusion (WBF)~\cite{solovyev2021weighted} to improve localization and confidence reliability.

We apply the \textit{Adaptive Labelling} method (see Section~\ref{subsubsection: AdaptiveLabelling}) using the OC-SORT tracker, configured with a detection confidence threshold of $det\_thresh = 0.3$, an Intersection-over-Union (IoU) threshold for association of $iou\_threshold = 0.85$, and a maximum age of $max\_age = 10$ frames to handle temporary object occlusions. These settings are designed to enhance the reliability and continuity of object tracks.

During the refinement stage, we adopt a base detection confidence threshold $\theta_0 = 0.3$, an adjustment factor $\alpha = 0.35$, a track quality threshold $\theta_q = 0.4$, and a penalty factor $\lambda = 0.1$. These parameters collectively govern the adaptive update of label confidences, aiming to suppress unreliable detections while preserving consistent track identities.







\subsection{Dataset}

We conducted experiments on a combined test set from the AI City Challenge 2023 and 2024~\cite{aicity}. As ground-truth annotations for the test videos were not publicly provided, we manually re-annotated them using the nine-class schema defined in the 2024 training set. As summarized in \autoref{tab:Table3}, the annotations include bounding boxes for motorcycles and each rider (driver or passenger as seen in \autoref{fig:rider_position}), with labels indicating helmet usage status. All models were trained solely on the publicly available 2024 training set.

\begin{table}[t!]
\centering
\caption{Ablation study on VisionGuard components. AL and CE denote the Adaptive Labeling and Contextual Expander modules, respectively. Results are reported in terms of mAP@50 (\%).}
\label{tab:results1}
\resizebox{\linewidth}{!}{
\begin{tabular}{l|c|c|l|l}
\toprule
\multirow{2}{*}{\textbf{Method}} & \multicolumn{2}{c|}{\textbf{Components}} 
  & \multicolumn{2}{c}{\textbf{Transformer Model}} \\ \cmidrule{2-5}
& \textbf{AL} & \textbf{CE}  & \textbf{Co-DETRs} & \textbf{DETR} \\ \midrule
Baseline &   &    & 44.221 & 41.473 \\ \midrule
VisionGuard & $\checkmark$ & & 44.222 (+0.001\%) & 41.474 (+0.001\%) \\ \midrule
VisionGuard & $\checkmark$ & $\checkmark$ & \textbf{44.945 (+1.6\%)} & \textbf{42.760 (+3.1\%)} \\ \hline
\end{tabular}
}
\end{table}

\begin{table}[t!]
\centering
\caption{Per-class performance with and without VisionGuard. Results are reported as AP@50 (\%). Note: P2Helmet and P0Helmet classes are absent from the test set.}
\label{tab:results2}
\begin{tabular}{c|cc|cc}
\toprule
\multirow{2}{*}{\textbf{Classes}} 
  & \multicolumn{2}{c|}{\textbf{Co-DETRs}} 
  & \multicolumn{2}{c}{\textbf{DETR}} \\ \cmidrule{2-5}
  & \textbf{Without} & \textbf{With} 
  & \textbf{Without} & \textbf{With} \\ \midrule
motorbike & 83.675 & \textbf{84.547} & 83.783 & \textbf{84.543}  \\ \midrule
DHelmet & 81.223 & \textbf{81.883} & 79.453 & \textbf{80.797} \\ \midrule
DNoHelmet & 78.294 & \textbf{79.231} & 72.754 & \textbf{74.499} \\ \midrule
P1Helmet & \textbf{4.004} & 3.922 & 0.000 & \textbf{0.117} \\ \midrule
P1NoHelmet & 62.356 & \textbf{64.701} & 54.326 & \textbf{59.037} \\  \midrule
P2NoHelmet & 0.000 & \textbf{0.011} & 0.000 & \textbf{0.047} \\ \midrule
P0NoHelmet & 0.000 & \textbf{0.317} & 0.000 & \textbf{0.283}  \\ \bottomrule
\end{tabular}
\end{table}


\subsection{Results}

To assess the effectiveness of our proposed post-processing framework, VisionGuard, on transformer-based object detectors, we conducted an ablation study using two representative architectures: DETR and Co-DETRs. Starting from the baseline detectors, we incrementally incorporated each component of VisionGuard, specifically Adaptive Labeling and Contextual Expander, to evaluate their individual and combined contributions, as detailed in \autoref{tab:results1}.

{Tables \ref{tab:results1} and \ref{tab:results2}} report consistent improvements across most categories after applying the refinement strategies of VisionGuard. The full method improves mAP@50 by {+3.1\%} on Co-DETR and {+1.6\%} on DETR, demonstrating its effectiveness.

Although the tracking-assisted Adaptive Labeling module provides limited benefits in some cases, this is likely due to the close spatial proximity and synchronized motion of riders and motorbikes, which degrades track quality and limits the efficacy of label corrections.

The Contextual Expander module, on the other hand, shows significant improvements, especially for underrepresented or contextually related categories. As shown in \autoref{tab:results2}, classes like \texttt{P1NoHelmet} show notable gains. Additionally, rare classes such as \texttt{P0NoHelmet} and \texttt{P2NoHelmet} benefit from the insertion of synthetic bounding boxes, resulting in small but meaningful improvements. Importantly, common categories such as \texttt{Motorbike} and \texttt{DHelmet} also show increased precision, confirming that VisionGuard does not disrupt the performance of high-confidence detections.

Overall, VisionGuard improves class coverage and detection precision by simulating plausible co-occurrence relationships and adjusting confidence scores, leading to a higher mAP without introducing additional false positives.

\section{Conclusion}

We introduced VisionGuard, a post-processing framework designed to improve the robustness and consistency of transformer-based object detectors for identifying helmet-use violations among motorcyclists. The framework incorporates two key modules: Adaptive Labeling, which resolves temporal inconsistencies across video frames, and Contextual Expander, which addresses class imbalance by injecting context-aware virtual bounding boxes with calibrated confidence scores. VisionGuard enhances detection precision, particularly for underrepresented and contextually relevant classes, while maintaining the integrity of high-confidence predictions. 

Future work will aim to improve helmet-use violation detection by incorporating finer-grained classification of helmet types and improper use (e.g., not fastening the straps), which are currently challenging for existing models. Moreover, extending the framework to handle multi-camera setups could improve coverage and reduce blind spots in large-scale surveillance systems.

\section*{Acknowledgments}

This research is supported by research funding from Faculty of Information Technology, University of Science, Vietnam National University - Ho Chi Minh City.

\printbibliography

\end{document}